\documentstyle[aclap,tree-dvips]{article}

\newtheorem{Theorem}{Theorem}
\newtheorem{Proposition}[Theorem]{Proposition}
\newtheorem{Lemma}[Theorem]{Lemma}
\newtheorem{Definition}[Theorem]{Definition}

\def\llimp{\mathrel{-\!\circ}}
\def\lltensor{\otimes}

\newcommand{\A}{{\cal A}}
\newcommand{\Bas}{\mbox{$\cal B$}}
\newcommand{\F}{\mbox{$\cal F$}}
\newcommand{\LMBK}{\mbox{\bf\sf L}}
\newcommand{\NL}{\mbox{\bf\sf NL}}
\newcommand{\NLP}{\mbox{\bf\sf NLP}}
\newcommand{\SDL}{\mbox{\bf\sf SDL}}
\newcommand{\SDLM}{\mbox{\bf\sf SDL$^-$}}
\newcommand{\LMBKfn}{\mbox{\footnotesize\bf\sf L}}
\newcommand{\SDLfn}{\mbox{\footnotesize\bf\sf SDL}}
\newcommand{\LP}{\mbox{\bf\sf LP}}
\newcommand{\tup}[1]{\langle #1 \rangle}

\newcommand{\seq}{\Rightarrow}

\newcommand{\proof}{{\bf Proof.}}
\newcommand{\myendproof}{\hfill $\Box$}

\newcommand{\lprove}{\vdash_{\LMBKfn}}
\newcommand{\sdlprove}{\vdash_{\SDLfn}}

\newcommand{\bsl}{\backslash}

\newcommand{\rname}[1]{(#1)}

\newcommand{\unarule}[3]{{\renewcommand{\arraystretch}{.9}
      \begin{array}{c}
         #2 \\ \hline #1
      \end{array}\;\rname{#3}
    }}
\newcommand{\unaruledisp}[3]{
  {\displaystyle {#2} \over {#1}}\hspace{1pt}{\rlap{$\scriptstyle #3$}}}

\newcommand{\binarule}[4]{{\renewcommand{\arraystretch}{.9}
      \begin{array}{cc}
         #2 \; &\; #3 \\ \hline \multicolumn{2}{c}{#1}
      \end{array}\;\rname{#4}
    }}

\newcommand{\binatree}[3]{\binatreepos{b}{#1}{#2}{#3}}

\newcounter{maxnode} 
\newcounter{nodea}
\newcounter{nodeb}
\newcounter{nodec}
\newcounter{noded}
\newcounter{nodee}
\newcounter{nodef}
\newcounter{nodeg}
\newcounter{nodeh}
\newcounter{nodei}
\newcounter{nodej}
\newcounter{numberofnodes}   
\newcommand{\newnodenames}{
  \addtocounter{nodea}{\thenumberofnodes}
  \addtocounter{nodeb}{\thenumberofnodes}
  \addtocounter{nodec}{\thenumberofnodes}
  \addtocounter{noded}{\thenumberofnodes}
  \addtocounter{nodee}{\thenumberofnodes}
  \addtocounter{nodef}{\thenumberofnodes}
  \addtocounter{nodeg}{\thenumberofnodes}
  \addtocounter{nodeh}{\thenumberofnodes}
  \addtocounter{nodei}{\thenumberofnodes}
}

\newcommand{\oldnodenames}{
  \addtocounter{nodea}{-\thenumberofnodes}
  \addtocounter{nodeb}{-\thenumberofnodes}
  \addtocounter{nodec}{-\thenumberofnodes}
  \addtocounter{noded}{-\thenumberofnodes}
  \addtocounter{nodee}{-\thenumberofnodes}
  \addtocounter{nodef}{-\thenumberofnodes}
  \addtocounter{nodeg}{-\thenumberofnodes}
  \addtocounter{nodeh}{-\thenumberofnodes}
  \addtocounter{nodei}{-\thenumberofnodes}
                           }

\newcommand{\binatreepos}[4]{
  \newnodenames   
  \begin{array}[#1]{cc}
    \node
      {\thenodeb}
      {$        #3
      $}
    &
   \node
      {\thenodec}
      {$
        \addtocounter{nodea}{\themaxnode}
        \addtocounter{nodeb}{\themaxnode}
        \addtocounter{nodec}{\themaxnode}
        #4
        \addtocounter{nodea}{-\themaxnode}
        \addtocounter{nodeb}{-\themaxnode}
        \addtocounter{nodec}{-\themaxnode}
      $}
   \\
   \\
   \multicolumn{2}{c}{
     \node
       {\thenodea}
       {$#2$}}
  \end{array}
  \nodeconnect[t]{\thenodea}[b]{\thenodeb}
  \nodeconnect[t]{\thenodea}[b]{\thenodec}
  \oldnodenames    }

\newcommand{\unatree}[2]{
      \begin{array}[b]{c}
         #2
         \\
         |
         \\
         #1
      \end{array}
   }

\newcommand{\lunatree}[2]{
      \begin{array}[b]{c}
         \mbox{{\tt #2}}
         \\
         #1
      \end{array}
   }

\title{Parsing for Semidirectional Lambek Grammar is NP-Complete}
\author{Jochen D\"orre\\[2ex]
  Institut f\"ur maschinelle Sprachverarbeitung\\
  University of Stuttgart}
\begin{document}
\maketitle

\begin{abstract}
  We study the computational complexity of the parsing problem
  of a variant of Lambek Categorial Grammar that we call {\em
    semidirectional}. In semidirectional Lambek calculus $\SDL$ there
  is an additional non-directional abstraction rule allowing the formula
  abstracted over to appear anywhere in the premise sequent's
  left-hand side, thus
  permitting non-peripheral extraction.
  $\SDL$ grammars are able to generate each context-free language and
  more than that. We show that the parsing problem for semidirectional
  Lambek Grammar is NP-complete by a reduction of the 3-Partition
  problem.\\[3ex]
\end{abstract}

\section{Introduction}
Categorial Grammar (CG) and in particular Lambek Categorial Grammar
(LCG) have their well-known benefits for the formal treatment of
natural language syntax and semantics. The most outstanding of these
benefits is probably the fact that the specific way, how the complete
grammar is encoded, namely in terms of `combinatory potentials' of its
words, gives us at the same time recipes for the construction of
meanings, once the words have been combined with others to form larger
linguistic entities. Although both frameworks are equivalent in weak
generative capacity --- both derive exactly the context-free
languages ---, LCG is superior to CG in that it can cope in a 
natural way with extraction and unbounded dependency phenomena. For
instance, no special category assignments need to be stipulated to
handle a relative clause containing a trace, because it is analyzed,
via  hypothetical reasoning, like a traceless clause with the trace
being the hypothesis to be discharged when combined with the relative
pronoun.
\begin{figure*}[ht]
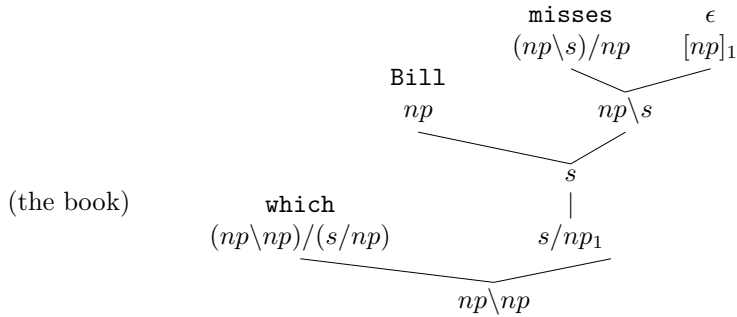

  \begin{displaymath}
    \vspace*{-\abovedisplayskip}
    \raisebox{13mm}{(the book)}\qquad
      {\binatree
        {np\bsl np}
        {\lunatree
           {(np\bsl np)/(s/np)}
           {which}
        }
        {\hspace*{-3em}
         \unatree
           {s/np_{1}}
           {\binatree
              {s}
              {\lunatree
                 {np}
                 {Bill}
              }
              {\binatree
                 {np\bsl s}
                 {\lunatree
                       {(np\bsl s)/np }
                       {misses}
                 }
                 {\lunatree
                    {{[np]_{1}}}
                    {$\epsilon$}
                 }
               }
             }
          }
       }
  \vspace*{\abovedisplayskip}
  \end{displaymath}
\caption{Extraction as resource-conscious hypothetical reasoning}
\label{fig:extr}
\end{figure*}
 Figure~\ref{fig:extr} illustrates this proof-logical
behaviour. Notice that this natural-deduction-style proof in the type
logic corresponds very closely to the phrase-structure tree one
would like to adopt in an analysis with traces. We thus can derive
{\tt Bill misses $\epsilon$} as an $s$ from the hypothesis that there
is a ``phantom'' $np$ in the place of the trace. Discharging the
hypothesis, indicated by index 1, results in {\tt Bill misses} being
analyzed as an $s/np$ from zero hypotheses. Observe, however, that
such a bottom-up synthesis of a new unsaturated type is only required,
if that type is to be consumed (as the antecedent of an
implication) by another type. Otherwise there would be a simpler proof
without this abstraction. In our example the relative pronoun has such
a complex type triggering an extraction.

A drawback of the pure Lambek Calculus \LMBK{} is
that it only allows for so-called `peripheral extraction', i.e., in
our example the trace should better be initial or final in the
relative clause. 

This inflexibility of Lambek Calculus is one of the reasons why many
researchers study richer systems today.  
For instance, the recent work by Moortgat \cite{Moortgat94} gives a
systematic in-depth study of mixed Lambek systems, which integrate the
systems \LMBK, \NL, \NLP, and \LP. These ingredient systems are
obtained by varying the Lambek calculus along two dimensions: adding
the permutation rule (P) and/or dropping the assumption that the type
combinator (which forms the sequences the systems talk about) is
associative (N for {\em non-associative}).

Taken for themselves these variants of \LMBK{} are of little use in
linguistic descriptions. But in Moortgat's mixed system all the
different resource management modes of the different systems are left
intact in the combination and can be exploited in different parts of
the grammar. The relative pronoun {\tt which} would, for instance,
receive category $(np\bsl np)/(np\llimp s)$ with $\llimp$ being
implication in \LP,%
\footnote{The Lambek calculus with permutation \LP{} is also called
the ``nondirectional Lambek calculus'' \cite{vBenthem88}. In it the
leftward and rightward implication collapse.}
i.e., it requires as an argument ``an $s$ lacking an $np$
somewhere''.%
\footnote{Morrill \cite{Morrill94} achieves the same effect with a
  permutation modality $\triangle$ applied to the $np$ gap:
  $(s/\triangle np)$}.

The present paper studies the computational complexity of a variant of
the Lambek Calculus that lies between \LMBK{} and \LP,
the Semidirectional Lambek Calculus $\SDL$.%
\footnote{This name was coined by Esther K\"onig-Baumer, who employs a
  variant of this calculus in her LexGram system \cite{LexGram95} for
  practical grammar development.}
Since \LP{} derivability is known to be NP-complete, it is interesting
to study restrictions on the use of the \LP{} operator $\llimp$. A
restriction that leaves its proposed linguistic applications intact is
to admit a type $B\llimp A$ only as the argument type in functional
applications, but never as the functor. Stated prove-theoretically
for Gentzen-style systems, this amounts to disallowing the left rule
for $\llimp$. Surprisingly, the resulting system $\SDL$ can be stated
without the need for structural rules, i.e., as a monolithic system
with just one structural connective, because the ability of the
abstracted-over formula to permute can be directly encoded in the
right rule for $\llimp$.
\footnote{ It should be pointed out
 that the resource management in this calculus is very closely
 related to the handling and interaction of local valency and
 unbounded dependencies in HPSG.
 The latter being handled with set-valued features {\sc slash}, {\sc
   que} and {\sc rel} essentially emulates the permutation
 potential of abstracted categories in semidirectional Lambek
 Grammar. A more detailed analysis of the relation between HPSG and
 $\SDL$ is given in \cite{LexGram95}.}

Note that
our purpose for studying \SDL{} is not that it might be in any sense
better suited for a theory of grammar (except perhaps, because of its
simplicity), but rather, because it exhibits a core of logical
behaviour that any richer system also needs to include, at least if it
should allow for non-peripheral extraction. The sources of complexity
uncovered here are thus a forteriori present in all these richer
systems as well.

\section{Semidirectional Lambek Grammar}

\subsection{Lambek calculus}
The semidirectional Lambek calculus (henceforth SDL) is a variant of
J.\ Lambek's original \cite{Lambek58} calculus of syntactic types. We
start by defining the Lambek calculus and extend it to obtain SDL.

Formulae (also called ``syntactic types'') are built 
from a set of propositional variables (or ``primitive types'') $\Bas =
\{b_1,b_2,\ldots\}$ and the three binary connectives $\bullet$,
$\backslash$, $/$, called {\em product, left implication}, and {\em
  right implication}. We use generally capital letters $A$, $B$, $C$,
\ldots to denote formulae and capitals towards the end of the alphabet
$T$, $U$, $V$, \ldots to denote sequences of formulae. The
concatenation of sequences $U$ and $V$ is denoted by $(U,V)$.

The (usual) formal framework of these logics is a Gentzen-style
sequent calculus. Sequents are pairs $(U,A)$, written as $U\seq
A$, where $A$ is a type and $U$ is a sequence of types.%
\footnote{In contrast to Linear Logic \cite{Girard87} the order of
  types in $U$ is essential, since the structural rule of permutation
  is not assumed to hold. Moreover, the fact that only a single
  formula may appear on the right of $\seq$, make the Lambek calculus
  an intuitionistic fragment of the multiplicative fragment of
  non-commutative propositional Linear Logic.}
The claim embodied by sequent $U\seq A$ can be read as ``formula $A$
  is derivable from the structured database $U$''.
Figure~\ref{fig:Lambek calc} shows Lambek's original calculus \LMBK. 

\begin{figure*}[thb]
\[
\begin{array}{c}
\unarule{b\seq b}{}{Ax}\\[2em]
\binarule{U, A/B, T, V\seq C}{T\seq B}{U,A,V\seq C}{/L}
\quad\quad\quad
\unarule{U\seq A/B}{U, B\seq A}{/R}\; \mbox{ if $U$ nonempty}\\[2em]
\binarule{U, T, B\backslash A, V\seq C}{T\seq B}{U, A, V\seq
      C}{\backslash L}
\quad\quad\quad
\unarule{U\seq B\backslash A}{B, U\seq A}{\backslash R}\; \mbox{ if $U$
  nonempty}\\[2em] 
\unarule{U, A\bullet B, V\seq C}{U, A, B, V\seq
      C}{\bullet L}
\quad\quad\quad
\binarule{U, V\seq A\bullet B}{U\seq A}{V\seq B}{\bullet R}\\[2em]
\binarule{U, T, V\seq C}{T\seq A}{U, A, V\seq C}{Cut}
\end{array}
\]
\caption{Lambek calculus \LMBK}
\label{fig:Lambek calc}
\end{figure*}

First of all, since we don't need products to obtain our results and
since they only complicate matters, we eliminate products from
consideration in the sequel.

In Semidirectional Lambek Calculus we add as additional connective the
\LP{} implication $\llimp$, but equip it only with a right rule.
\[
\unarule{T\seq B\llimp A}{U, B, V\seq A}{\llimp R}\; \mbox{ if $T=(U,V)$
  nonempty.} 
\]

Let us define the polarity of a subformula of a sequent $A_1,\ldots,A_n\seq A$
as follows: $A$ has positive polarity, each of $A_i$ have negative
polarity and if $B/C$ or $C\bsl B$ has polarity $p$, then $B$ also has
polarity $p$ and $C$ has the opposite polarity of $p$ in the sequent.

A consequence of only allowing the $\rname{\llimp R}$ rule, which is
easily proved by induction, is that in any derivable sequent $\llimp$
may only appear in positive polarity. Hence, $\llimp$ may not occur in
the (cut) formula $A$ of a $\rname{Cut}$ application and any
subformula $B\llimp A$ which occurs somewhere in the prove must also
occur in the final sequent. When we assume the final sequent's RHS to
be primitive (or $\llimp$-less), then the $\rname{\llimp R}$ rule will
be used exactly once for each (positively) occuring
$\llimp$-subformula. In other words, $\rname{\llimp R}$ may only do
what it is supposed to do: extraction, and we can directly read off
the category assignment which extractions there will be. 

We can show Cut Elimination for this calculus by a straight-forward
adaptation of the Cut elimination proof for \LMBK{}. We omit the proof
for reasons of space.
\begin{Proposition}[Cut Elimination]
Each SDL-derivable sequent has a cut-free proof.
\end{Proposition}

The cut-free system enjoys, as usual for Lambek-like logics, the {\em
  Subformula Property}: in any proof only subformulae of the goal
sequent may appear.

\newcommand{\hsh}{\#}

In our considerations below we will make heavy use of the well-known
count invariant for Lambek systems \cite{vBenthem88}, which is an
expression of the resource-consciousness of these logics. 
Define $\hsh_b(A)$ ({\em the $b$-count of $A$}), a function counting
positive and negative occurrences of primitive type $b$ in an
arbitrary type $A$, to be 
{
\[
\hsh_b(A) = 
\left\{ 
\begin{array}{@{ }l@{ }l@{}}
1 & \mbox{ if }A=b\\
0 & \mbox{ if $A$ primitive and }A\not=b\\
\hsh_b(B)-\hsh_b(C) & \mbox{ if }
\begin{array}[t]{@{}l@{}}
A=B/C\mbox{ or }A=C\bsl B\\
\mbox{ or }A=C\llimp B
\end{array}
\\
\hsh_b(B)+\hsh_b(C) & \mbox{ if }A=B\bullet C
\end{array}
\right.
\]
}

The invariant now states that for any primitive $b$, the $b$-count of
the RHS and the LHS of any derivable sequent are the same. By noticing
that this invariant is true for \rname{Ax} and is preserved by the
rules, we immediately can state:

\begin{Proposition}[Count Invariant]
If $\sdlprove U\seq A$, then $\hsh_b(U)=\hsh_b(A)$ for any $b\in\Bas$.
\end{Proposition}


Let us in parallel to \SDL{} consider the fragment of it in which
$\rname{/R}$ and $\rname{\bsl R}$ are disallowed. We call this
fragment \SDLM{}. Remarkable about this fragment is that any positive
occurrence of an implication must be $\llimp$ and any negative one
must be $/$ or $\bsl$.

\begin{figure*}[htb]
\[
\unaruledisp
{A_1^n,\, A_2,\, B_1^n,\, B_2,\, C_1^n,\, C_2\seq x}{
 x\seq x \quad\quad
  {\unaruledisp
    {A_1^{n-1},\, A_2,\, B_1^n,\, B_2,\, C_1^n,\, C_2\seq c\llimp(b\llimp x)}
    {\unaruledisp
      {\begin{array}{@{}c@{}}
          A_2,\, B_1^n,\, B_2,\, C_1^n,\, C_2,\, c^n,\, b^n\seq x\\
          \vdots\\
          A_1^{n-1},\, A_2,\, B_1^n,\, B_2,\, C_1^n,\, C_2,\, c,\,
          b\seq x
        \end{array}
      }
      {x\seq x \quad\quad
        {\unaruledisp
          {B_1^n,\, B_2,\, C_1^n,\, C_2,\, c^n,\, b^n\seq c\llimp(b\llimp y)}
          {\begin{array}{@{}c@{}}
              \vdots\\
              B_1^n,\, B_2,\, C_1^n,\, C_2,\, c^{n+1},\, b^{n+1}\seq
              y\;\;(\star)
            \end{array}
          }
          {2\times\rname{\llimp R}}
      }}
      {\rname{/L}}
    }
    {2\times\rname{\llimp R}}
  }}{
 \rname{/L}}
\]
\caption{Proof of $A_1^n,\, A_2,\, B_1^n,\, B_2,\, C_1^n,\, C_2\seq x$}
\label{fig:proof}
\end{figure*}

\subsection{Lambek Grammar}

\begin{Definition}
We define a {\em Lambek grammar} to be a quadruple
$\tup{\Sigma,\F,b_S,l}$ consisting of the finite alphabet of
terminals $\Sigma$, the set $\F$ of all Lambek formulae generated from
some set of propositional variables which includes the distinguished
variable $s$, and the
lexical map $l:\Sigma\to 2^{\cal F}$ which maps each terminal to a
finite subset of $\F$.
\end{Definition}

We extend the lexical map $l$ to nonempty strings of terminals by
setting $l(w_1 w_2 \ldots w_n) := l(w_1) \times l(w_2) \times \ldots
\times l(w_n)$ for $w_1 w_2 \ldots w_n\in\Sigma^+$.

The {\em language generated by a Lambek grammar}
$G=\tup{\Sigma,\F,b_S,l}$ is defined as the set of all strings $w_1
w_2\ldots w_n\in\Sigma^+$ for which there exists a sequence of types
$U\in l(w_1 w_2 \ldots w_n)$ and $\lprove 
U\seq b_S$. We denote this language by $L(G)$.

An {\em SDL-grammar} is defined exactly like a Lambek grammar, except that
$\sdlprove$ replaces $\lprove$.

Given a grammar $G$ and a string $w=w_1 w_2 \ldots w_n$, the {\em
parsing (or recognition) problem} asks the 
question, whether $w$ is in $L(G)$.

It is not immediately obvious, how the generative capacity of
SDL-grammars relate to Lambek grammars or nondirectional Lambek
grammars (based on calculus \LP). Whereas Lambek grammars generate
exactly the context-free languages (modulo the missing empty word)
\cite{Pentus93}, the latter generate all permutation closures of
context-free languages \cite{vBenthem88}. This excludes many
context-free or even regular languages, but includes some
context-sensitive ones, e.g., the permutation closure of $a^n b^n
c^n$. 

Concerning \SDL{}, it is straightforward to show that all context-free
languages can be generated by SDL-grammars.

\begin{Proposition}
Every context-free language is generated by some SDL-grammar.
\end{Proposition}
\proof{}
We can use a the standard transformation of an arbitrary cfr. grammar 
$G=\tup{N,T,P,S}$ to a categorial grammar $G'$. Since $\llimp$ does
not appear in $G'$ each \SDL-proof of a lexical assignment must be
also an \LMBK-proof, i.e. exactly the same strings are judged grammatical
by \SDL{} as are judged by \LMBK.
\myendproof

Note that since 
the $\{\rname{Ax},\rname{/L},\rname{\bsl L}\}$ subset of $\LMBK$ already
accounts for the cfr. languages, this observation extends to \SDLM.

Moreover, some languages which are not context-free can also be
generated. 

{\bf Example.}
Consider the following grammar $G$ for the language $a^n b^n c^n$.
We use primitive types $\Bas=\{b,c,x,y,z\}$ and define the lexical map
for $\Sigma=\{a,b,c\}$ as follows:
\addtolength{\arraycolsep}{-2pt}
\begin{eqnarray*}
  l(a) & := & \{ 
  \begin{array}[t]{c}
    x/(c\llimp(b\llimp x)),\\[-.4ex]
    = A_1
  \end{array}
  \begin{array}[t]{c}
    x/(c\llimp(b\llimp y))\\[-.4ex]
    = A_2
  \end{array}
  \}\\
  l(b) & := & \{
  \begin{array}[t]{c}
  (y/b)/y,\\[-.4ex]
    = B_1
  \end{array}
  \begin{array}[t]{c}
    (y/b)/z\\[-.4ex]
    = B_2
  \end{array}
  \}\\
  l(c) & := & \{
  \begin{array}[t]{c}
    (z/c)/z,\\[-.4ex]
    = C_1
  \end{array}
  \begin{array}[t]{c}
    z/c\\[-.4ex]
    = C_2
  \end{array}
  \}.
\end{eqnarray*}
The distinguished primitive type is $x$. To simplify the
argumentation, we abbreviate types as indicated above.

Now, observe that a sequent $U\seq x$, where $U$ is the image of some
string over $\Sigma$, only then may have balanced primitive counts, if
$U$ contains exactly one occurrence of each of $A_2$, $B_2$ and $C_2$
(accounting for the one supernumerary $x$ and balanced $y$ and $z$
counts) and for some number $n\geq 0$, $n$ occurrences of each of
$A_1$, $B_1$, and $C_1$ (because, resource-oriented speaking, each
$B_i$ and $C_i$ ``consume'' a $b$ and $c$, resp., and each $A_i$
``provides'' a pair $b$, $c$). Hence, only strings containing the same
number of $a$'s, $b$'s and $c$'s may be produced. Furthermore, due to
the Subformula Property we know that in a cut-free proof of $U\seq x$,
the main formula in abstractions (right rules) may only be either
$c\llimp(b\llimp X)$ or 
$b\llimp X$, where $X\in\{x,y\}$, since all other implication types
have primitive antecedents. 
Hence, the LHS of any sequent in the proof must be a subsequence of
$U$, with some additional $b$ types and $c$ types interspersed. But
then it is easy to show that $U$ can only be of the form
\[ A_1^n,\, A_2,\, B_1^n,\, B_2,\, C_1^n,\, C_2, \]
since any $/$ connective in $U$ needs to be introduced via
$\rname{/L}$. 

It remains to be shown, that there is actually a proof for such a
sequent. It is given in Figure~\ref{fig:proof}.

The sequent marked with $\star$ is easily seen to be derivable without
abstractions.

A remarkable point about $\SDL$'s ability to cover this language is
that neither $\LMBK$ nor $\LP$ can generate it. Hence, this example
substantiates the claim made in \cite{Moortgat94} that the inferential
capacity of mixed Lambek systems may be greater than the sum of its
component parts. Moreover, the attentive reader will have noticed that
our encoding also extends to languages having more groups of $n$
symbols, i.e., to languages of the form $a_1^n a_2^n\ldots a_k^n$. 

Finally, we note in passing that for this grammar the rules
$\rname{/R}$ and $\rname{\bsl R}$ are irrelevant, i.e. that it is at
the same time an \SDLM{} grammar.

\section{NP-Completeness of the Parsing Problem}

We show that the Parsing Problem for SDL-grammars is NP-complete by a
reduction of the 3-Partition Problem to it.%
\footnote{A similar reduction has been used in \cite{LincolnWinkler94} to show
  that derivability in the multiplicative fragment of propositional
  Linear Logic with only the connectives $\llimp$ and $\lltensor$
  (equivalently Lambek calculus with permutation $\LP$) is
  NP-complete.} This well-known
NP-complete problem is cited in \cite{np-completeness} as follows.
\begin{center}
  \begin{tabular}{l@{\hspace{2ex}}p{37ex}}
    Instance: 
    & Set $\A$ of $3m$ elements, a bound $N\in Z^+$, and a size
    $s(a)\in Z^+$ for each $a\in {\A}$ such that $\frac{N}{4} < s(a) <
    \frac{N}{2}$ and $\sum_{a\in {\A}} s(a)= mN$.\\
    Question:
    & Can $\A$ be partitioned into $m$ disjoint sets $\A_1,
    \A_2,\ldots,\A_m$ such that, for $1\leq i \leq m$, $\sum_{a\in \A_i}
    s(a) = N$ (note that each $\A_i$ must therefore contain exactly 3
    elements from $\A$)?\\
    Comment:
    & NP-complete in the strong sense.
  \end{tabular}
\end{center}

Here is our reduction. Let $\Gamma=\tup{\A,m,N,s}$ be a given 3-Partition
instance. For notational convenience we abbreviate
$(\ldots((A/B_1)/B_2)/\ldots)/B_n$ by $A/B_n\bullet
\ldots\bullet B_2\bullet B_1$ and similarly $B_n\llimp(\ldots
(B_1\llimp A)\ldots)$ by $B_n\bullet
\ldots\bullet B_2\bullet B_1\llimp A$, but note that this is just an
abbreviation in the product-free fragment. Moreover the notation $A^k$
stands for
\[ 
\begin{array}{c}
\underbrace{A\bullet A\bullet\ldots\bullet A}\\
k\mbox{ times}
\end{array}
\]
We then define the SDL-grammar $G_{\Gamma}=\tup{\Sigma,\F,b_S,l}$ as
follows: 
\[
\begin{array}{@{}l@{}}
  \Sigma  :=  \{v,w_1,\ldots,w_{3m}\}\\
  \F  :=  \begin{array}[t]{@{}l@{}}
    \mbox{\em all formulae over primitive types
  }\\
   \Bas=\{a,d\}\cup\bigcup_{i=1}^m \{b_i,c_i\}
 \end{array}
 \\
  b_S  :=  a\\
  l(v)  :=  a/(b_1^3\bullet b_2^3\bullet\ldots\bullet
  b_m^3\bullet c_1^N\bullet c_2^N\bullet\ldots\bullet c_m^N\llimp d)\\[1ex]
  \mbox{for $1\leq i\leq 3m-1$: }\\
  l(w_i)  :=  \bigcup_{1\leq j\leq m} d/d\bullet b_j\bullet
c_j^{s(a_i)}\\ 
  l(w_{3m})  :=  \bigcup_{1\leq j\leq m} d/b_j\bullet c_j^{s(a_{3m})}
\end{array}
\]

The word we are interested in is $v\, w_1\, w_2\, \ldots w_{3m}$. We
do not care about other words that might be generated by $G_{\Gamma}$.
Our claim now is that a given 3-Partition problem $\Gamma$ is solvable
{\bf if and only if} $v\, w_1 \ldots w_{3m}$ is in $L(G_{\Gamma})$.
We consider each direction in turn.

\begin{Lemma}[Soundness]
If a 3-Partition problem $\Gamma=\tup{\A,m,N,s}$ has a solution, then $v\,
w_1 \ldots w_{3m}$ is in $L(G_{\Gamma})$. 
\end{Lemma}
\proof{}
We have to show, when given a solution to $\Gamma$, how to choose a
type sequence $U\in l(v w_1 \ldots w_{3m})$ and construct an SDL proof
for $U\seq a$. Suppose $\A=\{a_1,a_2,\ldots,a_{3m}\}$. From a given
solution (set of triples) $\A_1, \A_2, \ldots, \A_m$ we can compute in
polynomial time a mapping $k$ that sends the index of an element to
the index of its solution triple, i.e., $k(i)=j$ iff $a_i\in \A_j$. To
obtain the required sequence $U$, we simply choose for the $w_i$
terminals the type $d/d\bullet b_{k(i)}\bullet c_{k(i)}^{s(a_i)}$
(resp. $d/b_{k(3m)}\bullet c_{k(3m)}^{s(a_{3m})}$ for $w_{3m}$).
Hence the complete sequent to solve is:
\begin{eqnarray*}
& & a/(b_1^3\bullet b_2^3\bullet\ldots\bullet
  b_m^3\bullet c_1^N\bullet c_2^N\bullet\ldots\bullet c_m^N\llimp d)\\
& & d/d\bullet b_{k(1)}\bullet c_{k(1)}^{s(a_1)}\\
& & \vdots\\
(*)
& & d/d\bullet b_{k(3m-1)}\bullet c_{k(3m-1)}^{s(a_{3m-1})}\\
& & d/b_{k(3m)}\bullet c_{k(3m)}^{s(a_{3m})}\\
& &\seq\\
& & a
\end{eqnarray*}

Let $a/B_0, B_1,\ldots B_{3m}\seq a$ be a shorthand for~$(*)$, and let
$X$ stand for the sequence of primitive types 
\[ b_{k(3m)},c_{k(3m)}^{s(a_{3m})}, b_{k(3m-1)},
c_{k(3m-1)}^{s(a_{3m-1})},\ldots b_{k(1)}, c_{k(1)}^{s(a_1)}.
\]
Using rule $\rname{/L}$ only, we can obviously prove $B_1,\ldots
B_{3m},X\seq d$. 
Now, applying $\rname{\llimp R}$ $3m+Nm$ times we can obtain
$B_1,\ldots B_{3m}\seq B_0$,
since there are in total, for each $i$, 3 $b_i$ and $N$ $c_i$ in $X$.
As final step we have
\[ \binarule{a/B_0,B_1,\ldots B_{3m}\seq a}{
             B_1,\ldots B_{3m}\seq B_0}{a\seq a}{/L}
\]
which completes the proof.
\myendproof

\begin{Lemma}[Completeness]
Let $\Gamma=\tup{\A,m,N,s}$ be an arbitrary 3-Partition problem and
$G_{\Gamma}$ the corresponding SDL-grammar as defined above. Then
$\Gamma$ has a solution, if $v\, w_1 \ldots w_{3m}$ is in
$L(G_{\Gamma})$.  
\end{Lemma}
\proof{}
Let $v\, w_1 \ldots w_{3m}\in L(G_{\Gamma})$ and 
\[
a/(b_1^3\bullet \ldots\bullet
  b_m^3\bullet c_1^N\bullet \ldots\bullet c_m^N\llimp d),
B_1,\ldots B_{3m}\seq a
\]
be a witnessing derivable sequent, i.e., for $1\leq i\leq 3m$, $B_i\in
l(w_i)$. Now, since the counts of this sequent must be balanced, the
sequence $B_1,\ldots B_{3m}$ must contain for each 
$1\leq j\leq m$ exactly 3 $b_j$ and exactly $N$ $c_j$ as subformulae.
Therefore we 
can read off the solution to $\Gamma$ from this sequent by including
in $\A_j$ (for $1\leq j\leq m$) those three $a_i$ for which $B_i$
has an occurrence of $b_j$, say these are $a_{j(1)}$, $a_{j(2)}$ and
$a_{j(3)}$. We verify, again via balancedness of the primitive counts,
that $s(a_{j(1)})+s(a_{j(2)})+s(a_{j(3)})=N$ holds, 
because these are the numbers of positive and negative occurrences of
$c_j$ in the sequent. This completes the proof.
\myendproof

The reduction above proves NP-hardness of the parsing problem. We need
strong NP-completeness of 3-Partition here, since our 
reduction uses a unary encoding. Moreover, the
parsing problem also lies within NP, since for a given grammar $G$
proofs are linearly bound by the length of the string and hence, we
can simply guess a proof and check it in polynomial time. Therefore we
can state the following:

\begin{Theorem}
The parsing problem for \SDL{} is NP-complete.
\end{Theorem}

Finally, we observe that for this reduction  the rules
$\rname{/R}$ and $\rname{\bsl R}$ are again irrelevant and that we can
extend this result to \SDLM.

\section{Conclusion}

We have defined a variant of Lambek's original calculus of types that
allows  abstracted-over categories to freely permute. Grammars based
on \SDL{} can generate any context-free language and more than that.
The parsing problem for \SDL{}, 
however, we have shown to be NP-complete. This result indicates that
efficient parsing for grammars that allow for large numbers of
unbounded dependencies from within one node may be problematic, even
in the categorial framework. 
Note that the fact, that this problematic case doesn't show up in
the correct analysis of normal NL sentences, doesn't mean that a
parser wouldn't have to try it, unless some arbitrary bound to that
number is assumed. 
For practical grammar engineering one can devise the motto {\em avoid
  accumulation of unbounded dependencies by whatever means}.

On the theoretical side we think that this result for \SDL{} is
also of some importance, since \SDL{} exhibits a core of
logical behaviour that any (Lambek-based) logic must have which accounts
for non-peripheral extraction by some form of permutation. And hence,
this result increases our understanding of the necessary computational
properties of such richer systems. To
  our knowledge the question, whether the Lambek 
  calculus itself or its associated parsing problem are NP-hard, are
  still open.


\end{document}